\pdfoutput=1

\documentclass[11pt]{article}

\usepackage{EMNLP2022}

\usepackage{times}
\usepackage{latexsym}
\usepackage{amsmath}
\usepackage{cleveref}
\usepackage{booktabs}
\usepackage{makecell}
\usepackage{multirow}
\usepackage{graphicx}

\usepackage[T1]{fontenc}

\usepackage[utf8]{inputenc}

\usepackage{microtype}

\usepackage{inconsolata}

\usepackage[normalem]{ulem}

\newcommand{\dataset}{{\scshape ProNCI}}
\newcommand{\clsgen}{{\scshape MtGen}}
\newcommand{\unigen}{{\scshape UniGen}}

\def\intitle{``\textit{Covid vaccine} is against Covid but \textit{Oxford vaccine} is made at Oxford!'' \\ Semantic Interpretation of Proper Noun Compounds}
\title{\intitle}

\author{Keshav Kolluru\textsuperscript{$\dagger$} \qquad Gabriel Stanovsky\textsuperscript{$\ddagger$} \qquad  Mausam\textsuperscript{$\dagger$} \\
  \textsuperscript{$\dagger$} Indian Institute of Technology Delhi\\
  \texttt{keshav.kolluru@gmail.com, mausam@cse.iitd.ac.in} \\
  \textsuperscript{$\ddagger$} The Hebrew University of Jerusalem\\
  \texttt{gabriel.stanovsky@mail.huji.ac.il}
}

\begin{document}
\maketitle

\begin{abstract}
\emph{Proper noun compounds}, e.g., ``\textit{Covid vaccine}'', convey information in a succinct manner (a ``Covid vaccine'' is a ``vaccine that \textit{immunizes against} the Covid \textit{disease}''). 
These are commonly used in short-form domains, such as news headlines, but are largely ignored in information-seeking applications.
To address this limitation, we release a new manually annotated dataset, \dataset{}, consisting of 22.5K proper noun compounds along with their free-form semantic interpretations. 
\dataset{} is 60 times larger than prior noun compound datasets  and also includes non-compositional examples, which have not been previously explored.
We experiment with various neural models for automatically generating the semantic interpretations from proper noun compounds, ranging from few-shot prompting to supervised learning, with varying degrees of knowledge about the constituent nouns.
We find that adding targeted knowledge, particularly about the common noun, results in performance gains of upto 2.8\%. 
Finally, we integrate our model generated interpretations with
an existing Open IE system and observe an 7.5\% increase in yield at a precision of 85\%.
The dataset and code are available at \href{https://github.com/dair-iitd/pronci}{https://github.com/dair-iitd/pronci}.

\end{abstract}

\section{Introduction}
\begin{table*}
\centering
\small
\begin{tabular}{lll}
\toprule
\textbf{Type} & \textbf{Example} & \textbf{Semantic Interpretations} \\ \midrule
Proper NC & \textit{Shakespeare biography} & is a biography about Shakespeare \\
(\textit{Proper-Common}) & \textit{London theatre} & is a theatre in London \textit{;} is a theatre located in London \\ 
& \textit{Concorde airplane} & \textsc{[Non-Cmp]} (Non-Compositional) \\
& \textit{Notre-Dame cathedral} & \textsc{[Non-Cmp]} (Non-Compositional) \\ \midrule
Common NC & \textit{nursing job} & is a job in nursing field \textit{;} is a job involving nursing \\ 
(\textit{Common-Common}) &  \textit{oil price} & is price paid for the oil\\ 
\bottomrule
\end{tabular}
\caption{Examples of common and proper noun compounds along with their semantic interpretations (``\textit{;}'' separates multiple interpretations). 
\textsc{[Non-Cmp]} indicates the absence of implicit relation between the constituent nouns.
}
\label{tab:intro}
\end{table*}

\emph{Proper noun compounds} (PNCs) \cite{breban19specialissue}\footnote{also referred to as proper noun modified compounds.} are grammatical constructions where a proper noun is followed by a common noun, for example: \textit{Covid vaccines} or \textit{Buddhist monks}.
These often serve as a compact way to convey information about an already known entity, omitting predicates that are interpreted by the reader using surrounding context, common sense, and world knowledge. For example, a reader is likely to interpret that ``Buddhist monks'' are ``\emph{religious people who are} buddhists''. In other cases, PNCs are used to identify specific entities, and do not provide additional information. For example, \textit{Watergate scandal} and \textit{Kawasaki disease} do not have any implicit relation between the proper and common noun 
as they refer to a specific instance of a scandal and a disease.
\Cref{tab:intro} provides additional examples. 

Thanks to their brevity, PNCs are commonly used to shorten descriptions in space-constrained domains, such as news articles headlines~\cite{breban19specialissue}. 
However, we find that prior work on compound noun interpretations only considered cases where the constituents are common nouns (e.g. \textit{baby oil}), thus missing all of the information conveyed in proper noun compounds~\cite{shwartz18paraphrase,hendrickx19semeval13t4}.

To address this limitation in current systems, we begin by defining the task of PNC interpretation as two subsequent stages~(\Cref{sec:definition}). 
The first stage requires identifying  whether a given PNC is compositional 
or not, while the second stage is the generation of an interpretation, where applicable.

In \Cref{sec:dataset}, we present \dataset{}, a crowd-sourced  dataset over Wikipedia containing 22.5K proper noun compounds and their annotated semantic interpretations. Candidates PNCs are found using syntactic parsing, and are then presented to crowdworkers who are asked to interpret them. Our annotation interface marks whether workers needed to read the full sentence, thus identifying PNCs whose interpretation relies on context. We will make the \dataset{} dataset publicly available to spur future research into PNCs.

In \Cref{sec:models}, we develop two approaches for PNC interpretation: (1) a multi-task neural model that performs classification and sequence generation in two distinct stages
and (2) a text-to-text approach, using a sequence-to-sequence model for both classification and generation.
In addition, we experiment with different methods for injecting various sources of world knowledge, which seems crucial for the task, using external resources like Wikipedia and WordNet \cite{fellbaum2010wordnet}, that give relevant information or definitions about the PNCs, that help in improving performance.  

For evaluating the generated interpretations, we propose a combination of classification-based metric and generation metrics  to properly handle both the interpretable and non-interpretable cases, respectively (\Cref{sec:setup}).
Since multiple correct interpretations are possible for a PNC, we use learned metrics such as BLEURT \cite{sellam2020bleurt}, that is finetuned on human-annotated preferences.

Finally, we show that training on \dataset{} yields models that can readily benefit extrinsic downstream application in the task of Open Information Extraction \cite{banko07oie}, thus widely extending their coverage (\Cref{sec:openie}).
Our approach first automatically extracts PNC interpretations using our models, then introduces it explicitly back into an Open IE extraction using a sequence to sequence model, thus giving an interpretation-integrated extraction. 
We then apply a high precision rule to generate new relations which leads to a 7.5\% increase in yield at an estimated precision of 85\% on the added extractions, when compared to extractions generated from the original sentences themselves. A major advantage of this approach is that it is agnostic to the Open IE system being used.
To conclude, our main contributions are:
\begin{enumerate}
    \item We introduce the \dataset{} dataset, containing interpretation for 22.5K proper noun compounds and their semantic interpretations. 
    \item We develop multi-task and generation based neural baselines that can leverage external knowledge for achieving higher performance.
    \item We design metrics for evaluating the quality of generated semantic interpretations.
    \item We demonstrate the usefulness of the generated interpretations in a downstream application by using them to augment the expressivity of Open IE systems.
\end{enumerate}



\section{Related Work}

Noun compounds are commonly used in English language, constituting 3.9\% of the tokens in the Reuters corpus \cite{baldwin04translation}.
They can be arbitrary length phrases, such as \textit{split air conditioner}, but most prior work on interpreting noun compounds has primarily looked at two word noun compounds of the type \textit{noun-noun}, where both are common nouns.
To the best of our knowledge challenges in interpretation where the first word is a proper noun (i.e., \textit{proper noun compounds}) have not been addressed, although their functional analysis and prevalence in certain domains have been studied in linguistics~\cite{rosenbach2007emerging,alexiadou2019proper,breban19specialissue}.
We briefly summarise the various types of noun-compound interpretations in literature and discuss their uses in applications.  

\paragraph{Types of interpretation:} Various types of interpretations for noun compounds have been explored, covering classification, ranking and generation.
Prior literature has frequently posed the interpretation as a \textbf{classification} task, where the classes can belong to abstract labels \cite{fares2016dataset},
semantic frame elements \cite{ponkiya18dataset} or prepositions \cite{lauer96statisticalnc}
However, none of these schemes can cover all range of possible noun compounds, thus limiting their expressivity and coverage.
SemEval 2010 Task 9 \cite{butnariu09semeval10t9} annotates human preferences for a set of 25-30 templatized paraphrases for each of the 250 training and 300 testing noun compounds. 
The task is framed as producing an accurate score for each paraphrase that \textbf{ranks} them in the correct order.
SemEval 2013 Task 4 \cite{hendrickx19semeval13t4} released a dataset of noun compounds and 
annotated free paraphrases for each compound.
Participating models were evaluated by matching and scoring the \textbf{generated} predictions with the gold set.

\citet{ponkiya20unsupervisednc} is the current state of art which poses the problem as generation of masked tokens using a pretrained T5 model \cite{raffel20t5} to get free paraphrase interpretations in a completely unsupervised manner. 
This leads to better performance than techniques that use the available training data. 
However, with the \dataset{} dataset, we do find that supervised models do outperform zero-shot models due to the scale.

\paragraph{Applications:} Noun compound interpretations have been helpful in translation of noun compounds  by either using a one-to-one mapping of interpreted prepositions \cite{paul10syntactic} or using recursive translation patterns \cite{balyan15translation}. 
In Question Answering systems, they have been used for disambiguating different types of noun-noun compounds in passage analysis \cite{ahn05qa}. They have also been useful for normalizing text that can help textual entailment \cite{nakov13entailment} and as auxiliary semantic annotation modules to improve parsing \cite{tratz11parsing}.
In this work, we show their use in the task of Open IE.

\paragraph{Open Information Extraction} (Open IE) \cite{banko07oie,mausam16survey,kolluru20imojie} involves extracting a set of tuples from the sentence where each field of the tuple contains phrases from the sentence itself. 
This makes it ontology-agnostic and allows it to be used for creation of domain agnostic Open Knowledge Bases \cite{broscheit20olpbench,vashishth18cesi,gupta19care}.
The relations are often verb-based \cite{fader11reverb} but can also be noun-mediated \cite{pal16relnoun} or involve implicit information \cite{soderland15implie}.

\citet{fader11reverb} relied on high precision rules to extract a wide variety of verb-mediated relations.
\citet{soderland15implie} uses dependency paths for generating high precision extractions based on three implicit relations, \textit{has job title}, \textit{has city} and \textit{has nationality}. 
\citet{pal16relnoun} considers noun mediated relations that can be extracted from compound noun phrases while dealing with challenges involved with denonyms and compound relational nouns. 
However, none of them consider implicit relations present in noun compounds. 

Moreover, recent state of art Open IE systems like OpenIE6 \cite{kolluru20openie6} and Gen2OIE \cite{kolluru22moie} rely on bootstrapped examples (generated using OpenIE4 \cite{pal16relnoun,christensen11oie4}) for training. 
Therefore they only generate extractions that contain phrases from the text and miss the cases where the content words are implicit. 
OpenIE6 \cite{kolluru20openie6} adopts a pipeline approach to integrate conjunction splitting into Open IE outputs, where coordination analysis and sentence splitting is performed as a preprocessing step, and the Open IE extractions are generated from the split sentences which are then merged.

\begin{table}[tb!]
\small
\centering
\begin{tabular}{p{7.0cm}}
\toprule
\textbf{Task Instructions} \\ \toprule
1. Your goal is to describe the relation between the two words by filling in the blanks. \\
2. You can write up to five words (or less!) 
3. The resulting relation should form a valid English sentence (see below for an example). \\
4. You can consult an example sentence as additional context, but the relation you write should be inferred only from the two words, and not by additional information. \\
5. If it is a name, entity, location or if you can't describe the relation between the words, please leave the relation blank. \\
\toprule
\textbf{Examples}     \\ \toprule
1. Coke Spokesman \textit{is a worker of} Coke. \\
2. Leake government \textit{is located in} Leake. \\
3. Capitol Hill \\
\toprule
\textbf{Pitfalls}     \\ \toprule
1. Coke Spokesman \textit{employment} Coke. \\
The relation should form a valid sentence. \\
2. Leake government \textit{has a failed} goverment.\\
The relation should be inferred by the words themselves and not by additional context. \\ \bottomrule
\end{tabular}
\caption{Instructions for the task along with examples and common pitfalls that are provided to the human workers from AMT for constructing \dataset{} dataset.}
\label{tab:annotation}
\end{table}

\section{Problem Definition}
\label{sec:definition}

Interpretations of noun compounds are meant to expose the expressed implicit relation.
Free-form paraphrases as interpretations provide flexibility for expressing 
relations implied in noun compounds, overcoming the limitations associated with choosing from a fixed set of classes or templates at the cost of a possibly non-consolidated representation, i.e., where similar-meaning noun compounds are represented differently.
Hence, we define semantic interpretation of a PNC as a free-form paraphrase that exposes the implicit relation between the constituent nouns, if any relation exists, else identify it as non-compositional (\textsc{[Non-Cmp]}). 

$$
\text{SemInt}(pnc)=
\begin{cases}
\text{Paraphrase}, & \text{if } \textit{reln.} \text{ exists}\\
\textsc{[Non-Cmp]}, & \text{if } \textit{reln.} \text{ absent}\\
\end{cases}
$$


\begin{figure*}[htp]
    \centering
    \includegraphics[width=15cm,height=3.5cm]{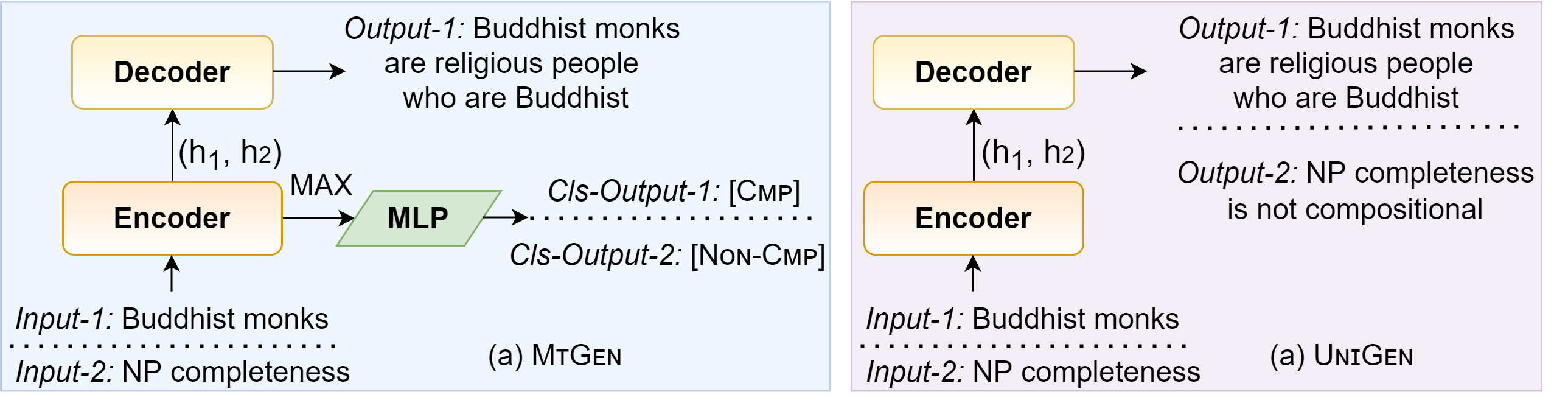}
    \caption{\clsgen{}, a multi-task Seq2Seq model classifies the example into (non) compositional classes and generates the interpretation where valid, while the \unigen{}, an unified generation model, uses a Seq2Seq model to generate interpretations or identify non-compositional examples using a specific string ``is not compositional''.}
    \label{fig:models}
\end{figure*}

\section{\dataset{} Dataset}
\label{sec:dataset}
To facilitate research in semantic understanding of proper noun compounds, we collect and release a supervised dataset as part of this work, which we call the \dataset{} dataset.
It contains 22,500 PNCs and their semantic interpretations which are annotated by human workers hired from Amazon Mechanical Turk (AMT).

The scale of the dataset is orders of magnitude greater than previously published free-paraphrase (common) noun compound datasets like SemEval 2013 Task 4~\cite{hendrickx19semeval13t4} that have only considered 355 noun compounds. 
For handling the evaluation of generated interpretations where multiple correct answers are possible, prior datasets choose to annotate multiple interpretations for each noun compound (varying from 30-50).
On the other hand, \dataset{} dataset only contains one interpretation per noun compound as we choose to invest our annotation budget in breadth rather than depth, relying on recent advances in semantic text similarity (e.g. BLEURT \cite{sellam2020bleurt}) to help evaluate the generated interpretations.

Moreover, prior datasets consider noun compounds out of  context, while \dataset{} also contains the sentence in which the proper noun compound is used. 
Providing this additional context helps to limit the ambiguity associated with multiple possible interpretations of the noun compound.
For example, \textit{U.S. sanctions} can mean either sanctions imposed by U.S. or sanctions imposed on U.S.
The exact case can be determined based on the context in which it is used.
``\textit{U.S. sanctions} on Iran have crippled the country'', implies the former and ``\textit{U.S. sanctions} by Iran...'' implies the latter.

To prepare the \dataset{} dataset, we randomly sample sentences from Wikipedia, and 
 retain sentences which contain two-word  proper noun compounds as identified by the SpaCy dependency parser \cite{spacy}. 
For every word, SpaCy identifies the root word along with the dependency tag. 
The ``compound'' dependency tag is used if the word and its root are part of a compound word. 
Then the parts of speech of the first and second word of the compound are checked. 
If they are proper noun (``PROPN'') and common noun (``NOUN'') respectively, we identify it as a proper noun compound and include it. 
If any word pairs have been identified incorrectly as proper noun compounds, they are marked by annotators to indicate the absence of any relation.

After the collection of proper noun compounds and corresponding sentences in which they appear, we posted HITs on the AMT platform for identification of relation between two words. 
The HITs were accompanied by task instructions, summarized in \Cref{tab:annotation}. 
The workers were paid 9 USD per hour on average, based on initial annotation experiments which indicated an average annotation time of 20 seconds on each compound.

To check the quality of annotation, we randomly sample 100 examples 
and find them to be correct 93\% of the time.
This represents an acceptable level, considering the difficulty of understanding certain compounds that need technical knowledge (\textit{AES key}) or cultural background (\textit{Abner characters}), as well as the subjectiveness in determining non-compositionality.

\begin{table*}[ht]
\small
\centering
\begin{tabular}{lp{12.5cm}}
\toprule
\textbf{Knowledge} & \textbf{Example} \\ \midrule
None & Buddhist monks \\
Sentence & Recent visitors to the campus include Buddhist monks who installed an environmental artwork at Lower Pond. {[}\textit{SEP}{]} Buddhist monks \\
WordNet-NN & Buddhist meaning: Buddhism is a widespread Asian religion based on a series of original teachings attributed to Gautama Buddha. {[}\textit{SEP}{]} Buddhist monks \\ 
Wiki-NNP & monks meaning: a male religious living in a cloister and devoting himself to contemplation and prayer and work {[}\textit{SEP}{]} Buddhist monks \\
NER-NNP & Buddhist belongs to nationalities or religious groups {[}\textit{SEP}{]} Buddhist monks \\ \bottomrule
\end{tabular}
\caption{Examples demonstrating the addition of different sources of knowledge for the compound, ``Buddhist monks'', in form of prompts that are concatenated with [\textit{SEP}] token. NNP and NN correspond for information about proper and common noun respectively, which can be from WordNet, Named Entity tags or Wikipedia.}
\label{tab:knowledge}
\end{table*}

\section{Models}
\label{sec:models}
The task of semantic interpretation of proper noun compounds involves generating valid paraphrases that explicate the relation in cases which are compositional.
So a model designed for this task needs to first identify if the given noun compound is compositional ([\textsc{Cmp}]) or not ([\textsc{Non-Cmp}]), and generate a paraphrase accordingly. 
We experiment with (1) supervised neural models, (2) adding external information and (3) zero/few-shot prompting models. 


\paragraph{Supervised neural models:} We use two types of supervised neural models: (1) a multi-task and (2) a unified generative model. 
Both models are depicted in \Cref{fig:models}.
The multi-task neural model uses a single model to perform both the tasks of classification as well as generation.
For classification, the model uses the max-pooled representations of encoder hidden states that is passed to an MLP \cite{maini20} to get the corresponding class probabilities of [\textsc{Cmp}] and [\textsc{Non-Cmp}]. 
In case the example is classified as compositional, a decoder is used for generating the paraphrase. 
We refer to this model as \clsgen{}.

In the unified generation model, we follow the recent advances in NLP where multiple tasks are posed in a common text-to-text format and are handled by a single Seq2Seq model like T5 \cite{raffel20t5}. 
For this purpose, we pose the task as a simple string generation problem that outputs either the paraphrase itself in cases where it is interpretable or generates the string ``\textit{proper noun compound} is non-compositional'' in the remaining cases.
We refer to this model as \unigen{}.

\paragraph{External information:} Since the task of interpretation requires knowledge of the noun compound, we also experiment with adding different types of knowledge to the model that help it in generating accurate interpretations.
Various methods have been proposed to incorporate external knowledge into pre-trained language models \cite{wang20kadapter,liu22relationalMALM,verga21film}. 
We use a simple strategy of concatenating the knowledge along with the proper noun compound before passing it to the model.
A [\textit{SEP}] token is added as a demarcator to differentiate the added knowledge.

We use four sources of knowledge that provide further information about the noun compound.
They include information of the proper noun, from (1) the first paragraph of Wikipedia that an entity linking system links it to (Wiki-NNP), (2) tags assigned to it by the Named Entity Recognition system (NER-NNP), or include information about the common noun using (3) the corresponding synset definitions provided in Wordnet (WordNet-NN), or information about the entire compound based on the (4) sentence in which it is used. 
An example of each type of knowledge is shown in \Cref{tab:knowledge}.

\paragraph{Zero/Few-shot prompting:}
Prior techniques for noun compound interpretation such as \cite{ponkiya20unsupervisednc} have proposed zero-shot generation using pre-trained language models to achieve state-of-art performance on SemEval 2013 Task 4 \cite{hendrickx19semeval13t4} and SemEval 2010 Task 9 \cite{butnariu09semeval10t9}.
We therefore evaluate the performance of such techniques along with some extensions using few-shot learning on the \dataset{} dataset. 
We find that there exists a significant gap compared to finetuning on the supervised dataset, demonstrating the importance of having a large scale dataset for the task of PNC interpretation.

\section{Experimental Setup}
\label{sec:setup}

The 22,500 examples of \dataset{} are split into train, validation and test such that all compounds with the same common noun occur exclusively in a single set.
Such splitting ensures that there is no intersecting common noun in either the train or evaluation splits.
This results in a more challenging setting than splitting the examples randomly, whose results are shown in \Cref{app:rand_split}.
Further, we also consider subsets that contain only compositional examples (CMP) or only non-compositional examples (Non-CMP).
The number of examples in each case are shown in \Cref{tab:dataset_splits}.

\begin{figure}[tb!]
    \centering
    \includegraphics[scale=0.35]{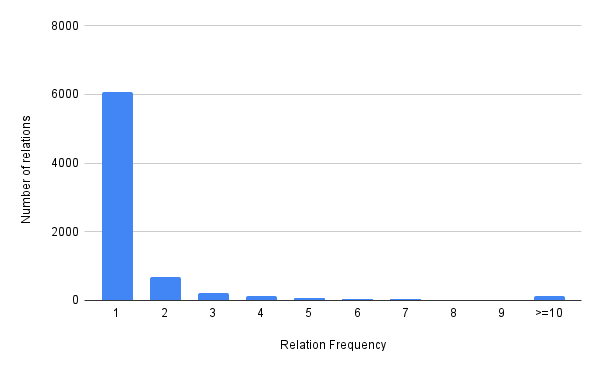}
    \caption{
    Plot of relation distribution in the \dataset{} dataset. 
    It shows the number of relations that have a frequency of 1 to 9 and $>=$10.}
    \label{fig:relation_dist}
\end{figure}

The dataset has 7,383 unique relations, with every relation occurring in an average of 1.84 examples.  
It contains 6,061 relations that occur only once in the dataset, as shown in \Cref{fig:relation_dist}.
The top 5 most commonly occurring relations along with their frequency (indicated in brackets) are \textit{is located in} (560), \textit{is based in} (389), \textit{are relatives of} (245), \textit{is an area of} (215) and \textit{are located in} (125). 


\begin{table}[tb!]
\small
\centering
\begin{tabular}{@{}lcccc@{}}
\toprule
Type & \#Train & \#Validation & \#Test & \#Total \\ \midrule
CMP & 9,722 & 1,416 & 2,497 & 14,389 \\
Non-CMP & 5,568 & 934 & 1,609 & 8,111 \\ \midrule
All & 15,290 & 2,350 & 4,106 & 22,500 \\ 
\bottomrule
\end{tabular}
\caption{
Number of training, validation and testing examples in the \dataset{} dataset. 
CMP indicates the subset that contains only compositional examples and constitute 63.9\% of the examples. 
Non-CMP indicates the complementary subset that contains only non-compositional examples and constitute the remaining 36.1\% of the examples.
}
\label{tab:dataset_splits}
\end{table}

\noindent \textbf{Evaluation metrics}:
Since the task involves a combination of classification and generation, the evaluation metric uses either exact match or semantic similarity depending on the type of example.
If an example has either the model prediction (\textbf{p}) or the gold annotation (\textbf{g}) as non-compositional, then an exact match (\textsc{Ex-Match}) between the prediction and gold gives a binary score of 0 or 1. 
In examples where both the gold annotation and model prediction are compositional, a semantic matching algorithm (\textsc{Sem-Match}) is used to give a score between 0 and 1 which indicates the extent of their similarity. 

$$\text{Score}(\textbf{g},\textbf{p}) = \begin{cases} \text{\textsc{Sem-Match}}(\textbf{g},\textbf{p}),\text{if CMP} \\
\text{\textsc{Ex-Match}}(\textbf{g},\textbf{p}),\text{if Non-CMP}. \\ \end{cases}$$

In particular, we compare two alternatives for \textsc{Sem-Match}: (1) the rule-based popular BLEU score, relies on \textit{n-gram} overlap, and often used in machine translations; and (2) BLEURT, which is finetuned over pretrained language model and represents a recent trend in trained evaluation metrics for text generation tasks.

In both alternatives, we use the entire paraphrase to evaluate the semantic score
as evaluating only the relations does not suit metrics such as BLEURT, which expects a full-formed sentence to infer the semantic meaning. 
Evaluation of the quality of BLUERT for similarity between predicted and gold paraphrases using 1K human annotated judgements indicates a 0.57 Pearson and 0.56 Kendall correlation.
We follow standard protocols in evaluating metric quality, as used in WMT Metrics shared tasks, 
and ask human annotators to rate the compositional model predictions as good, average and bad and see how these judgement scores correlate with the BLEU and BLEURT scores.
Further details are provided in \Cref{app:qa}.

We denote the final evaluation metric as \textsc{Sem/Ex-Match}.
When using BLEU or BLEURT as the semantic matcher, the metric is also referred to as \textsc{BLEU/Ex} or \textsc{BLEURT/Ex}, respectively.
To understand the effect of each type of match, we also report the \textsc{Ex-Match} classification accuracy over all the examples, where the compositional type is assigned the positive class, and the non-compositional type is assigned as the negative class.
Along with binary accuracy, we compute the precision and recall as well.
Since the \textsc{Sem-Match} cannot be computed over all examples, we report the scores averaged over only the cases where both gold and prediction are compositional.

\paragraph{Pre-trained models:} 
For all our experiments, we use the T5-base \cite{raffel20t5} as the default initialization, unless explicitly mentioned otherwise.
It contains 220M parameters.
For checking the statistical consistency, every model is trained 5 times with different seeds and their mean and standard deviation are reported.

\paragraph{Hyper-parameters and computational resources:} 
We run all our experiments using a V100 GPU.
We use the standard hyper-parameters recommended in T5 for all the experiments, using a batch size of 16, initial learning rate of 2e-5.
The final model is chosen using early stopping on the validation set after training for 10 epochs.
Each round of training and evaluation takes around 1 hr.

\begin{table*}[tb!]
\small
\centering
\begin{tabular}{@{}llrrrrrrrrrrr@{}}
\toprule
Model & Knowledge & \multicolumn{3}{c}{\textsc{Ex-Match}} & \multicolumn{2}{c}{\textsc{Sem-Match}} & \multicolumn{2}{c}{\textsc{Sem/Ex-Match}} \\ \cmidrule{3-9}
& & \multicolumn{1}{c}{Precision} & \multicolumn{1}{c}{Recall} & \multicolumn{1}{c}{Accuracy} & \multicolumn{1}{c}{BLEU} & \multicolumn{1}{c}{BLEURT} & \multicolumn{1}{c}{BLEU} & \multicolumn{1}{c}{BLEURT} \\ \midrule
\multirow{5}{*}{\clsgen{}} & None & \textbf{79.1} \tiny $\pm$ 1.37 & 67.1 \tiny $\pm$ 1.84 & \textbf{79.5} \tiny $\pm$ 0.58 & 32.7 \tiny $\pm$ 1.61 & 57.9 \tiny $\pm$ 0.42 & 44.3 \tiny $\pm$ 1.05 & 57.5 \tiny $\pm$ 0.66 \\
& Sentence & 78.1 \tiny $\pm$ 1.51 & 68.4 \tiny $\pm$ 2.50 & 79.4 \tiny $\pm$ 0.25 & 34.7 \tiny $\pm$ 0.36 & 58.3 \tiny $\pm$ 0.76 & 45.7 \tiny $\pm$ 0.60 & 57.8 \tiny $\pm$ 0.75 \\
& WordNet-NN & 74.2 \tiny $\pm$ 3.71 & 76.4 \tiny $\pm$ 5.68 & 79.4 \tiny $\pm$ 1.08 & 33.2 \tiny $\pm$ 1.08 & 57.6 \tiny $\pm$ 0.51 & 47.1 \tiny $\pm$ 0.92 & 58.9 \tiny $\pm$ 0.76 \\
& Wiki-NNP & 52.8 \tiny $\pm$ 2.43 & \textbf{90.6} \tiny $\pm$ 3.02 & 63.2 \tiny $\pm$ 2.96 & 24.0 \tiny $\pm$ 0.36 & 32.9 \tiny $\pm$ 2.38 & 43.0 \tiny $\pm$ 0.50 & 45.4 \tiny $\pm$ 0.98 \\
& NER-NNP & \textbf{79.1} \tiny $\pm$ 0.63 & 67.7 \tiny $\pm$ 1.63 & 79.7 \tiny $\pm$ 0.55 & 34.5 \tiny $\pm$ 0.23 & 59.2 \tiny $\pm$ 0.37 & 45.4 \tiny $\pm$ 0.51 & 58.3 \tiny $\pm$ 0.68 \\
\midrule
\multirow{5}{*}{\unigen{}} & None & 73.5 \tiny $\pm$ 2.99 & 74.4 \tiny $\pm$ 2.26 & 78.7 \tiny $\pm$ 1.40 & 34.1 \tiny $\pm$ 1.99 & 58.6 \tiny $\pm$ 0.78 & 46.7 \tiny $\pm$ 1.12 & 58.6 \tiny $\pm$ 0.94 \\
& Sentence & 73.0 \tiny $\pm$ 1.57 & 77.6 \tiny $\pm$ 1.83 & 79.3 \tiny $\pm$ 0.55 & 34.4 \tiny $\pm$ 0.81 & 58.8 \tiny $\pm$ 0.68 & \textbf{47.9} \tiny $\pm$ 0.41 & \textbf{59.5} \tiny $\pm$ 0.57 \\
& WordNet-NN & 65.3 \tiny $\pm$ 5.76 & 82.9 \tiny $\pm$ 5.05 & 74.5 \tiny $\pm$ 3.74 & 33.7 \tiny $\pm$ 0.88 & 56.5 \tiny $\pm$ 0.65 & 47.4 \tiny $\pm$ 0.45 & 56.7 \tiny $\pm$ 1.52 \\
& Wiki-NNP & 65.3 \tiny $\pm$ 3.05 & 66.3 \tiny $\pm$ 5.50 & 71.8 \tiny $\pm$ 1.32 & 25.7 \tiny $\pm$ 0.59 & 37.8 \tiny $\pm$ 2.13 & 38.4 \tiny $\pm$ 1.55 & 43.9 \tiny $\pm$ 1.09 \\
& NER-NNP & 75.7 \tiny $\pm$ 0.95 & 72.3 \tiny $\pm$ 1.52 & 79.4 \tiny $\pm$ 0.21 & \textbf{35.2} \tiny $\pm$ 0.23 & \textbf{59.4} \tiny $\pm$ 0.40 & 46.9 \tiny $\pm$ 0.45 & 59.0 \tiny $\pm$ 0.42 \\
\bottomrule
\end{tabular}
\caption{
Performance of \clsgen{} and \unigen{} on the \dataset{} dataset trained under five different knowledge settings.
All the models are evaluated using the three types of matching. 
`None' corresponds to using no external knowledge.
Adding external knowledge improves the performance of the models in three out of four cases.
}
\label{tab:sup_training}
\end{table*}



\begin{table*}[tb!]
\small
\centering
\begin{tabular}{@{}lcccccccc@{}}
\toprule
Model & \multicolumn{3}{c}{\textsc{Ex-Match}} & \multicolumn{2}{c}{\textsc{Sem-Match}} & \multicolumn{2}{c}{\textsc{Sem/Ex-Match}} \\ \cmidrule{2-8}
& \multicolumn{1}{c}{Precision} & \multicolumn{1}{c}{Recall} & \multicolumn{1}{c}{Accuracy} & \multicolumn{1}{c}{BLEU} & \multicolumn{1}{c}{BLEURT} & \multicolumn{1}{c}{BLEU} & \multicolumn{1}{c}{BLEURT} \\ \midrule
\citet{ponkiya20unsupervisednc} & 0.0 & 0.0 & 60.8 & 23.1 & 44.9 & 13.8 & 26.8 \\
Rand Few-Shot (5) & 37.3 & 11.0 & 55.3 & 27.7 & 40.2 & 18.5 & 25.1 \\
Rand Few-Shot (10) & 62.1 & 21.4 & 58.2 & 27.6 & 39.3 & 22.3 & 28.2 \\
KNN Few-Shot (5) & 68.7 & 43.6 & 69.1 & \textbf{29.9} & 46.1 & 33.1 & 41.4 \\
KNN Few-Shot (10) & \textbf{67.1} & \textbf{50.5} & \textbf{69.9} & \textbf{29.9} & \textbf{46.9} & \textbf{35.2} & \textbf{43.7} \\
\bottomrule
\end{tabular}
\caption{
Performance of T5 model without any finetuning.
\citet{ponkiya20unsupervisednc} corresponds to the zero-shot setting adapted from the corresponding paper.
Few-shot techniques use either five or ten example 
demonstrations.
In `Rand' the few-shot examples are chosen randomly while  in `KNN'  the nearest neighbours of the query are chosen as the few-shot examples.
Availability of annotated examples from \dataset{} helps to substantially improve the performance of the model.
Overall performance remains inferior to the finetuned models.
}
\label{tab:few_shot}
\end{table*}

\section{Experiments}
\label{sec:experiments}

In this section, we address the following questions:
\begin{enumerate}
    \item How do \unigen{} and \clsgen{} compare with each other and what benefit does adding external knowledge provide to these models?
    \item What is the performance difference between few-shot learning and supervised training? 
    \item How do individual components of the noun compound influence the model predictions? 
\end{enumerate}

\subsection{Performance of Supervised Models}
In \Cref{tab:sup_training}, we show the results of both, the multi-task model, \clsgen{} and the unified generation model, \unigen{} (\Cref{sec:models}).

We find that the \unigen{} model outperforms the \clsgen{} model in overall performance but leads to a modest drop in the compositionality classification performance.
For example, in the case where no additional knowledge is used, \unigen{} leads to a higher \textsc{Sem/Ex-Match} score with both BLEU and BLEURT leading to an increase of (2.4, 1.1) pts.
But \unigen{} achieves a lower classification score with the \textsc{Ex-Match} accuracy reducing by 0.8\%.
We attribute this observation to the fact that \clsgen{} uses a separate module that enables it to be tuned better for the classification task.
However, \unigen{} performs better in overall performance as both the encoder and decoder can benefit from positive transfer between the tasks.

By adding knowledge to the model, using the prompting described in \Cref{tab:knowledge}, at both training and testing time, we see gains in performance in three out of four types of knowledge added. 
Using information of the proper noun from Wikipedia often reduces the performance due to incorrect entity linking.
Among the the remaining three sources of knowledge, we find that WordNet-NN leads to the maximum increase in performance in three of the four settings.
We find that the predicted interpretations are often biased to re-use words that occur in the knowledge prompts and this leads to higher scores in case of less frequently occurring compounds.
For instance, the prediction changes from ``Kirati community is a group of Kirati'' to ``Kirati community are people of Kirati'', when added with the knowledge, ``Major groups of Kirati community follows Buddhism''.
Using student paired t-test we find that improvements are statistically significant with \textit{p}-value of 3.78$e^{-10}$ of BLEURT scores averaged over all 5 seeds.
We do not find additional improvements when multiple knowledge sources are added simultaneously (\Cref{sec:multiple_knowledge}). 

Predictions of \unigen{} trained with sentence knowledge are rated to be 72\% correct when checked manually on a sample of 100 sentences. 
This indicates a significant scope for improvement, when compared to the upper bound of 93\% data quality (\Cref{sec:dataset}).

We conduct two further experiments on the trained \unigen{} model to understand the strength of semantic matching used and the effectiveness of the model on the related task of common noun compound interpretation.

\paragraph{Template scoring:}
To test the effect of template word matching on BLEU and BLEURT scores, we introduce a dummy relation: i.e., the prediction for every non-compositional example is forced to be ‘noun-compound is none of common-noun’. This ensures that only template words match, but the semantic meaning is wrong. 
On re-computing the \textsc{Sem-Match} scores of \unigen{}, this reduces the BLEU score from 34.1 to 22.9 and BLEURT score from 46.7 to -3. 
This follows the expected trend as BLEU gives partial scores to template matches, but BLEURT focuses on the overall semantic meaning. 

\paragraph{SemEval evaluation:}
When \unigen{} is evaluated on the free noun compound paraphrasing task of SemEval 2013 Task 4 \citep{hendrickx19semeval13t4}, it achieves an isomorphic score of 72.8 compared to 80.1 reported by \citet{ponkiya18dataset}. 
We attribute this to different interpretation styles with \dataset{} focusing on detailed relations (average length of 6.9 words) compared to SemEval (average length of 5.1 words), leading to slightly lower scores with word match heuristics adopted by the task.

\subsection{Performance of few-shot learning}
State-of-art models for free paraphrase interpretations of \textit{common} noun compounds \cite{ponkiya20unsupervisednc} uses the zero-shot generation capabilities of T5 and show that they outperform supervised models.
To check if the same holds for the \dataset{} dataset, we experiment with zero-shot generation.
Similar to \cite{ponkiya20unsupervisednc}, we use the masked template, ``$w_1 w_2$ is a <\textit{extra}\_\textit{id}\_\textit{0}> the $w_1$'', where T5 fills in the missing words in place of <\textit{extra}\_\textit{id}\_\textit{0}>. 

We further experiment with few-shot learning, where \textit{K} training examples are chosen as part of the prompt which the model can use to perform in-context learning and generate the prediction for the given input.
No additional knowledge is used in these set of experiments.
These \textit{K} examples can either be chosen randomly or the nearest neighbours to the input query can be chosen, where cosine distance between the input and a training example is computed after embedding them with a pre-trained T5-Encoder \cite{liu22gpt3incontext}. 
We experiment with \textit{K} = 5 or 10.
The limitations of context size in the pretrained models prevent us from testing with higher values of \textit{K}.

In \Cref{tab:few_shot}, we find that the zero-shot performance trails behind the best few-shot model with a decrease of 21.4, 41 pts in BLEU/\textsc{Ex}, BLEURT/\textsc{Ex}, respectively.
This is partly because of the variety of examples in the \dataset{} dataset, which cannot be fit into specific templates and the inability of zero-shot model to handle non-compositional examples.
In few-shot learning, expanding the prompt size and dynamically choosing the prompt examples helps achieve higher performance but the performance still remains lower than the fully-supervised \unigen{} model which is still 11.2, 15.3 pts higher in BLEU/\textsc{Ex}, BLEURT/\textsc{Ex}.


\subsection{Proper noun vs. Common noun}

The interpretation of a proper noun compound depends on both the proper noun and common noun present in it.
To study how each of the two nouns influence the prediction, we randomly shuffle the their characters in both input and gold annotation.

In \Cref{tab:ind_components}, we find that common noun has a larger effect on the model performance as shuffling its characters leads to a significant drop performance of (5.8, 17.6, 35) pts in (BLEU/\textsc{Ex}, BLEURT/\textsc{Ex}, \textsc{Ex-Match} Accuracy\%). 
Comparatively, the proper noun results in a much smaller drop of (3.1, 8, 16.3) pts in the three evaluation metrics.




\begin{table}[tb!]
\small
\centering
\begin{tabular}{@{}lcrrrr@{}}
\toprule
Shuffle & \multicolumn{1}{c}{\textsc{Ex-Match}} & \multicolumn{2}{c}{\textsc{Sem/Ex-Match}}\\ \midrule
& \multicolumn{1}{c}{Accuracy} & \multicolumn{1}{c}{BLEU} & \multicolumn{1}{c}{BLEURT} \\ \cmidrule{2-4}
None & \textbf{78.7} \tiny $\pm$ 1.40 & \textbf{46.7} \tiny $\pm$ 1.12 & \textbf{58.6} \tiny $\pm$ 0.94 \\
NNP & 62.4 \tiny $\pm$ 0.97 & 43.6 \tiny $\pm$ 1.01 & 50.6 \tiny $\pm$ 0.44 \\
NN & 43.7 \tiny $\pm$ 1.02 & 40.9 \tiny $\pm$ 0.15 & 41.0 \tiny $\pm$ 0.16 \\ \bottomrule
\end{tabular}
\caption{\unigen{} evaluated after random shuffling of characters in the proper (NNP) or common (NN) noun.}
\label{tab:ind_components}
\end{table}

\section{Application to Open IE}
\label{sec:openie}
To demonstrate the downstream value of the noun compound interpretations, we add them to a state-of-art Open IE system, Gen2OIE \cite{kolluru22moie}, and generate new extractions that capture implicit relations.
We apply this on a corpus of 21,228 COVID-19 news headlines that contain proper noun compounds like COVID-19 outbreak, Rohingya refugee, etc \cite{aslam20headlines}. 

\noindent \textbf{Integration}: To achieve this, we train a Seq2Seq model that takes as input the sentence concatenated with the interpretation of the PNC present in it and outputs an interpretation-augmented sentence.
For example, the sentence, ``Workers sound alarm on \textit{Covid-19 outbreak}'' and the interpretation, ``\textit{Covid-19 outbreak} is an outbreak of Covid-19'' are integrated to get the following output, ``Workers sound alarm on outbreak of Covid-19''.
Considering the simplicity of the task, we annotate a small set of 200 examples of this kind and use it to train a Seq2Seq model.
Since this style of integration converts the implicit relation in the noun compound to an explicit form, it allows for the Open IE system to add new relations that were missing earlier.

\begin{figure}[tb!]
    \centering
    \includegraphics[width=0.9\columnwidth]{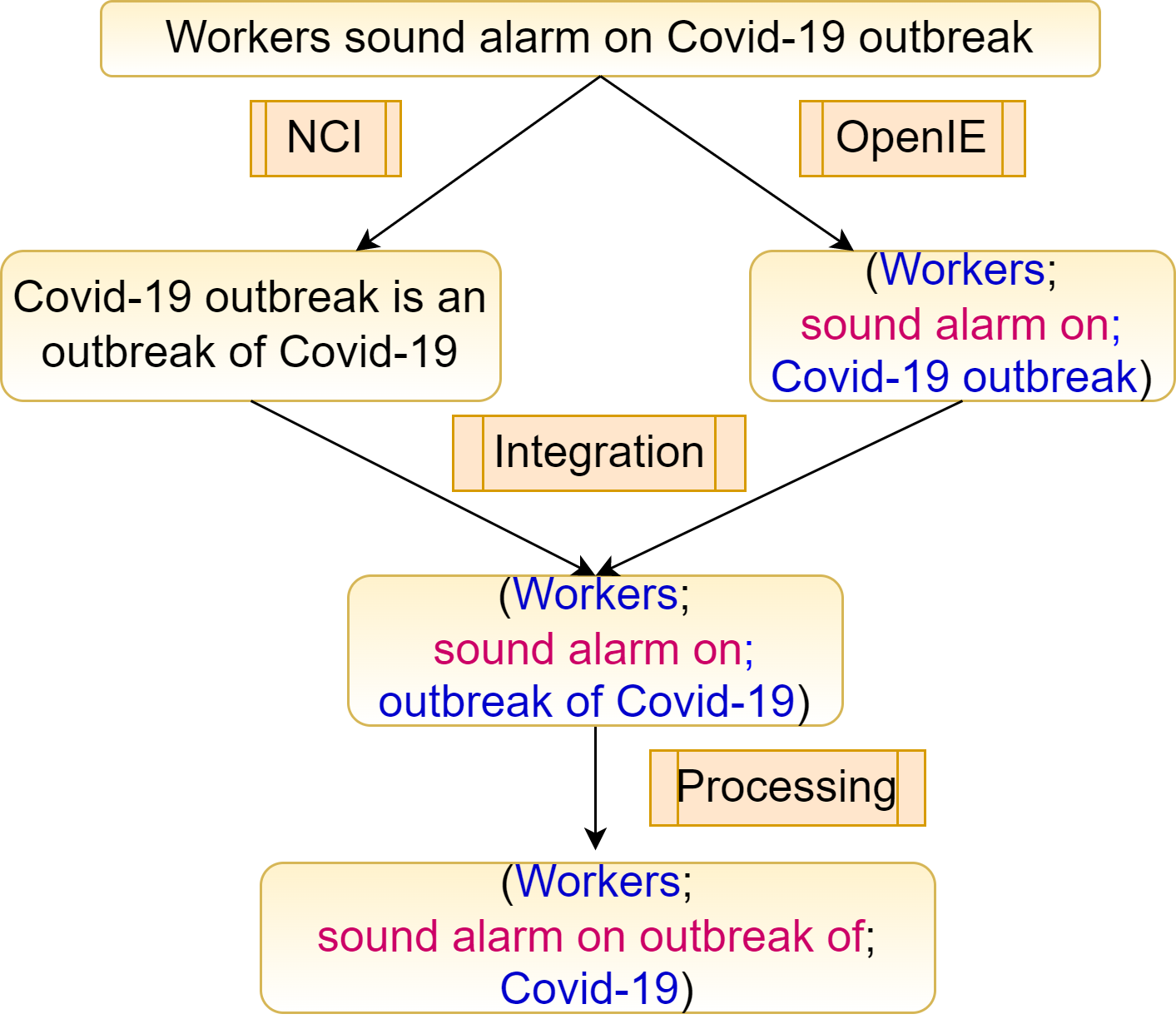}
    \caption{Open IE Pipeline. Postprocessing of the extraction integrated with noun compound interpretation, generates the new extraction.}
    \label{fig:openie_int}
\end{figure}

\noindent \textbf{Processing}: We experiment with a high precision rule that post-processes an extraction to generate a new one, whenever the extraction contains a PNC at the start of its object.  
For example, if the original extraction is (Workers; sound alarm on; COVID-19 outbreak), and the corresponding integrated extraction is (Workers; sound alarm on; outbreak of COVID-19), then the rule generates a new extraction by moving words till the proper noun back into the relation. 
In this case, we get the extraction, (Workers; sound alarm on outbreak of; COVID-19) -- thus exposing a direct relationship between workers and COVID-19, which was not present earlier.
The overall pipeline is shown in \Cref{fig:openie_int}.

We find that extractions generated using this pipeline leads to an increase in yield of 7.5\% where the added extractions have a precision of 85\%, compared to a precision of 82.2\% of the original extractions, as determined on a random sample of 500 extractions. 
We note that the method can use any Open IE system without any additional finetuning to produce the noun-compound extractions.


\section{Conclusion}
In this work we develop the novel task of semantic intepretation of proper noun compounds. 
We present the \dataset{} dataset for the task and test performance of various neural models.
We show the downstream utility of generated interpretations by integrating it with an Open IE system that results in generation of new extractions involving implicit relations.
Linguistic characterization of non-compositionality and utilizing additional sources of knowledge present scope for further work.

\section{Limitations}
The proposed models are evaluated on a specific test set, which may not be representative of all the types of examples that it may encounter during deployment.
Due to the use of pretrained models, the system may also exhibit possible biases that have been discovered in the pretrained models.

\section*{Acknowledgements}
Keshav is supported by a TCS Fellowship. 
Mausam is supported by grants from Huawei, Google, Verisk, and a Jai Gupta Chair Fellowship.
We thank KnowDis team for their help in data annotations and HPC, IIT Delhi for the computational resources.
This work was supported in part by a research grant no. 2088 from the Israeli Ministry of Science and Technology.

\bibliography{anthology,custom}
\bibliographystyle{acl_natbib}

\clearpage
\appendix
\twocolumn[\centering \Large\bfseries \intitle\\ \large (Appendix) \\ \vspace{2ex} ]

\section{Quality Assesment of Evaluation Metrics}
\label{app:qa}

For evaluating the quality of the metrics that are used for evaluating the model predictions, in particular, the semantic matching component \Cref{sec:setup}, we manually annotate the quality of model predictions with respect to gold using a 3-index scale.
The scale indicates whether the quality of the prediction is bad, average or good.
This is done only for the cases where the gold annotation indicates that the compound is compositional and the prediction of the model is also a paraphrase, as semantic matching is applicable only in these cases.
A total of 1500 examples are annotated out of which 500 are used for finetuning the learned metrics such as BLEURT.
On the remaining 1K examples, we compute the Pearson and Kendall correlation between the scores assigned by the evaluation metric and the human annotated scores.
We report the results in \Cref{tab:app_qa} for five evaluation metrics which include BLEU, BLEURT with and without finetuning on both the base and large variants. 
We find that the fine-tuned BLEURT outperforms both BLEU and the un-trained BLEURT.
It specifically outperforms BLEU by a significant margin from 0.28 to 0.57 in Pearson correlation and 0.23 to 0.46 in Kendall correlation. 
We find that the performance of both the base and large variants of BLEURT perform similarly after being finetuned and a minor difference exists in their untuned variants.

We note that the correlation of 0.57 is on par with the current state of NLG metrics. 
For example, \citet{chen20mocha}, reports a correlation of 0.45-0.60 for standard metrics such as BLEU, BERTScore \citep{bert-score} on short-text evaluation.
To further encourage research in building better generation metrics, we release the human judgements of the interpretations.

\begin{table}[tb!]
\small
\centering
\begin{tabular}{@{}lrr@{}}
\toprule
Metric & \makecell{Pearson \\ $|\rho|$} & \makecell{Kendall \\ $\tau$} \\ \midrule
BLEU & 0.28 & 0.23 \\
BLEURT-base & 0.43 & 0.37 \\
BLEURT-large & 0.49 & 0.4 \\
BLEURT-base \textit{(tuned)} & 0.56 & \textbf{0.46} \\
BLEURT-large \textit{(tuned)} & \textbf{0.57} & \textbf{0.46} \\ \bottomrule
\end{tabular}
\caption{Quality of metrics evaluated using Pearson and Kendall rank correlation. (tuned) indicates models that are fine-tuned on 500 manually evaluated comparisons.}
\label{tab:app_qa}
\end{table}

\section{Random Split of \dataset{}}
\label{app:rand_split}

In this section, we evaluate the results of \unigen{} and \clsgen{} on a random split of the \dataset{} dataset, where the 22,500 examples are randomly split into 17,500 training, 2,500 validation and 2,500 testing examples.
The results are reported in \Cref{tab:rand_sup_results} and \Cref{tab:rand_unsup_results}.
We find that the performance is higher compared when split according to common nouns.
This can be attributed to the lack of intersecting common nouns between the training and evaluation sets that could have provided additional clues. 
This leads to a drop in (BLEU/\textsc{Ex}, BLEURT/\textsc{Ex}, \textsc{Ex} Acc\%) scores of (5.7, 5.4, 3.3) pts in \clsgen{} and (5.3, 4.6, 2.9) pts in \unigen{}.

\begin{table*}[tb!]
\small
\centering
\begin{tabular}{@{}llrrrrrrrrrrr@{}}
\toprule
Model & Knowledge & \multicolumn{3}{c}{\textsc{Ex-Match}} & \multicolumn{2}{c}{\textsc{Sem-Match}} & \multicolumn{2}{c}{\textsc{Sem/Ex-Match}} \\ \cmidrule{3-9}
& & \multicolumn{1}{c}{Precision} & \multicolumn{1}{c}{Recall} & \multicolumn{1}{c}{Accuracy} & \multicolumn{1}{c}{BLEU} & \multicolumn{1}{c}{BLEURT} & \multicolumn{1}{c}{BLEU} & \multicolumn{1}{c}{BLEURT} \\ \midrule
\multirow{5}{*}{\clsgen{}} & None & \textbf{78.2} \tiny $\pm$ 1.14 & 74.5 \tiny $\pm$ 1.48 & 82.8 \tiny $\pm$ 0.25 & 40.5 \tiny $\pm$ 0.58 & 63.8 \tiny $\pm$ 0.36 & 50.0 \tiny $\pm$ 0.39 & 62.9 \tiny $\pm$ 0.29 \\
& Sentence & 76.9 \tiny $\pm$ 1.70 & 78.6 \tiny $\pm$ 1.38 & 83.3 \tiny $\pm$ 0.69 & 40.4 \tiny $\pm$ 0.43 & 63.2 \tiny $\pm$ 0.30 & 51.1 \tiny $\pm$ 0.25 & 63.4 \tiny $\pm$ 0.50 \\
& WordNet-NN & 76.4 \tiny $\pm$ 1.30 & 80.5 \tiny $\pm$ 1.95 & \textbf{83.5} \tiny $\pm$ 0.36 & 40.8 \tiny $\pm$ 0.63 & 63.3 \tiny $\pm$ 0.40 & 51.8 \tiny $\pm$ 0.55 & 63.8 \tiny $\pm$ 0.47 \\
& Wiki-NNP & 51.7 \tiny $\pm$ 1.04 & 94.7 \tiny $\pm$ 0.82 & 65.2 \tiny $\pm$ 1.41 & 25.9 \tiny $\pm$ 1.26 & 36.0 \tiny $\pm$ 3.80 & 42.9 \tiny $\pm$ 0.44 & 46.0 \tiny $\pm$ 1.14 \\
& NER-NNP & 75.4 \tiny $\pm$ 2.19 & 80.5 \tiny $\pm$ 3.06 & 82.9 \tiny $\pm$ 0.45 & 40.5 \tiny $\pm$ 0.79 & 63.4 \tiny $\pm$ 0.62 & 51.4 \tiny $\pm$ 0.29 & 63.5 \tiny $\pm$ 0.26 \\
\midrule
\multirow{5}{*}{\unigen{}} & None & 71.7 \tiny $\pm$ 0.68 & 83.4 \tiny $\pm$ 1.07 & 81.6 \tiny $\pm$ 0.21 & 41.5 \tiny $\pm$ 0.16 & 63.7 \tiny $\pm$ 0.17 & 52.0 \tiny $\pm$ 0.24 & 63.2 \tiny $\pm$ 0.15 \\
& Sentence & 72.1 \tiny $\pm$ 0.32 & 83.6 \tiny $\pm$ 0.44 & 81.9 \tiny $\pm$ 0.19 & 41.3 \tiny $\pm$ 0.19 & 63.4 \tiny $\pm$ 0.45 & 52.0 \tiny $\pm$ 0.12 & 63.3 \tiny $\pm$ 0.17 \\
& WordNet-NN & 71.0 \tiny $\pm$ 1.71 & \textbf{86.2} \tiny $\pm$ 1.23 & 81.7 \tiny $\pm$ 0.88 & \textbf{42.0} \tiny $\pm$ 0.40 & 64.0 \tiny $\pm$ 0.39 & \textbf{52.9} \tiny $\pm$ 0.34 & \textbf{63.8} \tiny $\pm$ 0.42 \\
& Wiki-NNP & 68.6 \tiny $\pm$ 2.08 & 68.2 \tiny $\pm$ 1.93 & 76.5 \tiny $\pm$ 0.82 & 26.1 \tiny $\pm$ 0.78 & 39.0 \tiny $\pm$ 2.18 & 38.7 \tiny $\pm$ 0.42 & 45.3 \tiny $\pm$ 1.25 \\
& NER-NNP & 71.9 \tiny $\pm$ 0.98 & 81.8 \tiny $\pm$ 1.70 & 81.3 \tiny $\pm$ 0.25 & 41.6 \tiny $\pm$ 0.34 & \textbf{64.2} \tiny $\pm$ 0.68 & 51.6 \tiny $\pm$ 0.42 & 63.1 \tiny $\pm$ 0.49 \\
\bottomrule
\end{tabular}
\caption{
Performance of the two models, \clsgen{} and \unigen{} on the randomly split \dataset{} dataset trained under five different knowledge settings.
}
\label{tab:rand_sup_results}
\end{table*}

\begin{table*}[tb!]
\small
\centering
\begin{tabular}{@{}lcccccccc@{}}
\toprule
Model & \multicolumn{3}{c}{\textsc{Ex-Match}} & \multicolumn{2}{c}{\textsc{Sem-Match}} & \multicolumn{2}{c}{\textsc{Sem/Ex-Match}} \\ \cmidrule{2-8}
& \multicolumn{1}{c}{Precision} & \multicolumn{1}{c}{Recall} & \multicolumn{1}{c}{Accuracy} & \multicolumn{1}{c}{BLEU} & \multicolumn{1}{c}{BLEURT} & \multicolumn{1}{c}{BLEU} & \multicolumn{1}{c}{BLEURT} \\ \midrule
\citet{ponkiya20unsupervisednc} & 0.0 & 0.0 & 62.8 & 22.9 & 44.1 & 14.4 & 27.7 \\
Rand Few-Shot (5) & 53.7 & 1.2 & 63.0 & 27.6 & 41.2 & 17.7 & 26.2 \\
Rand Few-Shot (10) & 37.7 & 33.6 & 54.4 & 28.8 & 42.2 & 24.7 & 29.7 \\
KNN Few-Shot (5) & \textbf{70.5} & 53.0 & 74.3 & 34.8 & 51.7 & 38.7 & 48.0 \\
KNN Few-Shot (10) & 68.4 & \textbf{60.1} & \textbf{74.8} & \textbf{35.4} & \textbf{53.2} & \textbf{41.0} & \textbf{50.3} \\
\bottomrule
\end{tabular}
\caption{Performance of T5 model without any finetuning on the random split of \dataset{} dataset.}
\label{tab:rand_unsup_results}
\end{table*}

\section{Effect of Pretraining}
\label{app:pretraining}
To understand the effect pretraining has on the effect of model performance for the task of semantic interpretation of proper noun compounds, we re-train the \unigen{} on the NOUN split starting from random initialization, instead of using T5-base, the default in all of our experiments. 
We also experiment with using T5-large.
We report the results in \Cref{tab:app_pretraining}.
We find that Random initialization is considerably worse, where the scores reduces from 46.7 to 33.9 in BLEU/EM and 58.6 to 30.5 in BLEURT/EM.
This indicates that pretrained initialization plays a significant role in the final performance on the task.
Moreover, on experimenting with the larger model, T5-large, we find a slight increase in scores from (46.7, 58.6, 78.7) to (47.7, 58.7, 79.4) in (BLEU/EM, BLEURT/EM, CMP).
Thus the task can benefit from scaling of the language models as they typically gain more information about the common and proper nouns.

\begin{table}[tb!]
\small
\centering
\begin{tabular}{@{}lcrr@{}}
\toprule
Init & \multicolumn{1}{c}{\textsc{Ex-Match}} & \multicolumn{2}{c}{\textsc{Sem/Ex-Match}}\\ \midrule
& \multicolumn{1}{c}{Accuracy} & \multicolumn{1}{c}{BLEU} & \multicolumn{1}{c}{BLEURT} \\ \cmidrule{2-4}
Random & 63.9 \tiny $\pm$ 1.98 & 33.9 \tiny $\pm$ 1.25 & 30.5 \tiny $\pm$ 1.27 \\
T5-base & 78.7 \tiny $\pm$ 1.40 & 46.7 \tiny $\pm$ 1.12 & 58.6 \tiny $\pm$ 0.94 \\
T5-large & \textbf{79.4} \tiny $\pm$ 0.11 & \textbf{47.7} \tiny $\pm$ 0.29 & \textbf{58.7} \tiny $\pm$ 0.35 \\ \bottomrule
\end{tabular}
\caption{Performance of the \unigen{} model on the \dataset{} dataset trained using different initializations of the Seq2Seq model.
Random initialization leads to huge drop in performance. 
}
\label{tab:app_pretraining}
\end{table}

\section{Error Analysis}
\label{app:error_analysis}
We analyze the mistakes made by the \unigen{} model trained with Sentence Knowledge to find potential scopes for improvement.
We divide them into the following categories - 
\begin{enumerate}
    \item Lack of word sense disambiguation: We notice mistakes in the model predictions in cases when some words have multiple meanings. The model defaults to choosing the one with most frequent usage and not disambiguating properly based on the context. 
    For example, the the interpretation, ``\textit{Sunday strip} is a comic printed on a Sunday'' is mistaken as ``\textit{Sunday strip} is a show on  Sunday'', even when the sentence contains sufficient clues for the same.
    The given sentence is ``In a few cases, the topper introduced characters later developed into a successful Sunday strip.''
    \item Non Informative predictions: Although predictions are not wrong they are often not very informative. For example, the model produces the following interpretation, ``\textit{EU economies} are based in EU'' compared to the more detailed gold ``\textit{EU economies} are the financial condition of EU members''.
    \item Errors in evaluation and mistakes in Gold: In some cases, the evaluation metric is unable to capture semantic similarity. For example, the model prediction ``Baltimore hospitals are located in Baltimore'' and the gold, ``Baltimore hospitals are medical institutions in Baltimore'', has a BLEURT score of only -0.11.
\end{enumerate}


\section{Adding multiple sources of knowledge}
\label{sec:multiple_knowledge}

\begin{table}[tb!]
\small
\centering
\begin{tabular}{lcrrrrr}
\toprule
Knowledge & \multicolumn{1}{c}{\textsc{Ex-Match}} & \multicolumn{2}{c}{\textsc{Sem/Ex-Match}}\\ \midrule
& \multicolumn{1}{c}{Accuracy} & \multicolumn{1}{c}{BLEU} & \multicolumn{1}{c}{BLEURT} \\ \cmidrule{2-4}
Sentence & 79.3 \tiny $\pm$ 0.55 & \textbf{47.9} \tiny $\pm$ 0.41 & \textbf{59.5} \tiny $\pm$ 0.57 \\
 +WNet-NN & 77.4 \tiny $\pm$ 2.14 & 46.5 \tiny $\pm$ 1.48 & 57.4 \tiny $\pm$ 1.63 \\
 +Wiki-NNP & 74.0 \tiny $\pm$ 1.62 & 38.9 \tiny $\pm$ 4.21 & 46.1 \tiny $\pm$ 6.38 \\
 +NER-NNP & \textbf{79.4} \tiny $\pm$ 0.23 & 47.0 \tiny $\pm$ 0.52 & 58.9 \tiny $\pm$ 0.40 \\ \bottomrule
\end{tabular}
\caption{Performance of the \unigen{} model on \dataset{} dataset trained with additional sources of knowledge added over Sentence knowledge.
The additional sources do not provide further benefits.}
\label{tab:app_additional_knowledge}
\end{table}

In \Cref{sec:experiments}, we observed statistically significant benefits to model performance after adding information about the noun compound from various sources of knowledge.
We also experiment with adding information from multiple source of knowledge to see if it can further augment the model performance.
On taking the best performing Sentence knowledge in the \unigen{} model on NOUN split, we add the remaining three sources of knowledge and report their performance in \Cref{tab:app_additional_knowledge}.
We find that it results in a slightly decrease in performance in case of WNet-NN and NER-NNP and in case of Wiki-NNP the decrease is much greater because of the reduced quality of Wikipedia entities.
We attribute this to possible confusion arising from disparate sources of knowledge that highlight different parts of the noun compound.

\end{document}